\documentclass[runningheads]{llncs}
\usepackage{multirow}
\usepackage{amsmath}
\usepackage{array}
\usepackage{amsfonts}
\usepackage{nccmath}
\usepackage{amssymb}
\usepackage{marvosym}
\usepackage[T1]{fontenc}
\usepackage{graphicx,verbatim}
\usepackage{caption}
\begin{document}

\title{ReSurgSAM2: Referring Segment Anything in Surgical Video via Credible Long-term Tracking}

\titlerunning{ReSurgSAM2: Referring Segment Anything in Surgical Video}

\author{
    Haofeng Liu\inst{1} \and
    Mingqi Gao\inst{2} \and
    Xuxiao Luo\inst{1}\ \and
    Ziyue Wang\inst{1} \and
    Guanyi Qin\inst{1} \and \\
    Junde Wu\inst{3} \and
    Yueming Jin\inst{1}\textsuperscript{(\Letter)}
}
\authorrunning{Liu et al.}

\institute{
    National University of Singapore, Singapore, Singapore \and
    Southern University of Science and Technology, Shenzhen, China \and
    University of Oxford, Oxford, United Kingdom \\
    \email{haofeng.liu@u.nus.edu}, \email{ymjin@nus.edu.sg}
}

\maketitle              
\begin{abstract}
Surgical scene segmentation is critical in computer-assisted surgery and is vital for enhancing surgical quality and patient outcomes. Recently, referring surgical segmentation is emerging, given its advantage of providing surgeons with an interactive experience to segment the target object. However, existing methods are limited by low efficiency and short-term tracking, hindering their applicability in complex real-world surgical scenarios.
In this paper, we introduce ReSurgSAM2, a two-stage surgical referring segmentation framework that leverages Segment Anything Model 2 to perform text-referred target detection, followed by tracking with reliable initial frame identification and diversity-driven long-term memory.
For the detection stage, we propose a cross-modal spatial-temporal Mamba to generate precise detection and segmentation results. 
Based on these results, our credible initial frame selection strategy identifies the reliable frame for the subsequent tracking.
Upon selecting the initial frame, our method transitions to the tracking stage, where it incorporates a diversity-driven memory mechanism that maintains a credible and diverse memory bank, ensuring consistent long-term tracking.
Extensive experiments demonstrate that ReSurgSAM2 achieves substantial improvements in accuracy and efficiency compared to existing methods, operating in real-time at 61.2 FPS. Our code and datasets will be available at https://github.com/jinlab-imvr/ReSurgSAM2.


\end{abstract}

\section{Introduction}
Surgical video scene segmentation is vital in computer-assisted surgery. Through precise identification and differentiation of surgical instruments and tissues, this technology can provide cognitive assistance to surgeons for decision-making~\cite{moglia2021systematic}.
Most existing methods in surgical scene segmentation rely only on video data~\cite{ref_Yue2024surgsam,zhou2023text}.
Despite achieving real-time and satisfactory performance, these systems can only generate semantic masks for all instruments and tissues collectively, without providing surgeons the ability to interactively identify and track specific objects of interest.
Referring segmentation, which aims to automatically identify and segment the specific object according to a textual expression, is beneficial for various surgical applications~\cite{zhou2023text,wang2024video}. Integrated with AR technology, it could enhance surgical education by enabling trainees to interactively explore specific instruments through AR overlays~\cite{kovoor2021validity,sheik2019next}. During intraoperative surgery, such ability enables the system to focus on regions of interest specified by surgeons, which can optimize workflow, providing accurate and personalized navigation support, contributing to safer and higher-quality outcomes of patient care.

To enable this capability, RSVIS~\cite{wang2024video} takes the first step to investigate referring instrument segmentation in the surgical domain. They propose a video-instrument synergistic network to learn both video-level and instrument-level knowledge to improve performance. However, this method relies only on short-term information of three consecutive frames, leading to inherent challenges in long-term tracking.
Referring segmentation in surgery still remains under-explored. 
In natural domains, the typical task is referring video object segmentation (RVOS). The online RVOS methods ~\cite{wu2023onlinerefer,li2024refsam} can offer real-time processing, however, they lack robust long-term tracking essential for surgical scenarios, as these methods are typically developed for short videos (under 10 seconds). This limitation is particularly critical in surgical procedures, which generally last long durations in hours, with dynamic scene variations and instrument movement.
In contrast, offline RVOS alternatives~\cite{wu2022language,yan2024referred} achieve better performance through extended temporal integration but require future frame information, making them unsuitable for intraoperative applications.

Recently, Segment Anything Model 2 (SAM2)~\cite{sam2} has gained attention for its interactive framework with satisfactory tracking capabilities, presenting a promising potential to enhance the referring segmentation task.
By providing visual prompts on initial frames as permanent memory, it can perform consistent tracking with memory attention. 
However, visual prompts (e.g., bounding boxes or multi-points) rely on the object's presence in the first few frames and impose labeling burdens on surgeons during operations. Since the target object may not be present at the beginning of surgery, such interactions are suboptimal~\cite{moglia2021systematic}. 
Instead, textual expressions offer greater flexibility and are the closest form to audio. Integrating textual prompts into SAM2 marks a crucial step toward hands-free interaction in surgery~\cite{surgicalsam2,wang2024video}.
Furthermore, identifying the reliable initial frame for tracking is critical for RVOS, as inaccurate segmentation masks can introduce error accumulation, significantly degrading overall performance~\cite{cuttano2024samwise}.
Finally, SAM2 employs a greedy strategy that selects the nearest frames as memory without assessing their reliability, limiting long-term temporal modeling~\cite{liu2022learning}.

To address these challenges, we propose ReSurgSAM2, a novel two-stage framework built on SAM2 that sequentially performs text-referred target detection followed by tracking, enabling efficient and accurate RVOS in long-range surgical videos.
During the detection stage, we identify the reliable initial frame using our Cross-Modal Spatial-Temporal Mamba (CSTMamba) and Credible Initial Frame Selection (CIFS) strategy. 
CSTMamba efficiently captures dedicated spatial-temporal dependencies across video frames while integrating multi-modal features, enabling precise specified object detection and segmentation for robust initial frame selection. 
Leveraging these accurate detection results, CIFS selects the optimal frame for tracking initialization based on confidence.
During tracking, our Diversity-Driven Long-term Memory (DLM) mechanism empowers SAM2 to track the object reliably throughout the entire surgical video by conditioning on a diverse and dependable memory bank. 
Extensive experiments on datasets containing instruments and tissues demonstrate significant performance improvements over existing methods, maintaining real-time at 61.2 FPS.

\section{Method}
\subsection{ReSurgSAM2 Framework}
SAM2~\cite{sam2} extends SAM~\cite{ref_kirillov2023sam} to video domains by incorporating temporal memory attention while preserving its segmentation capabilities.
It implements a short-term memory mechanism through a queue, which conditions the current frame features on both the permanent initial frame and nearest predictions through greedy selection.
For each frame, the mask decoder generates predictions with two key scores: the intersection-over-union (IoU) score and the occlusion score.
The IoU score estimates the alignment between the prediction with ground truth, while the occlusion score employs a signed confidence scheme - positive values signify object presence, negative values indicate absence, and the magnitude reflects confidence. 
This dual-scoring system enables object tracking even during occlusions.
While SAM2 demonstrates promising performance in general video domains, it faces limitations in adapting to surgical RVOS, including vision-language integration, reliable initial frame identification, and long-term tracking.

\begin{figure}[tbp]
    \centering
    \includegraphics[width=\textwidth]{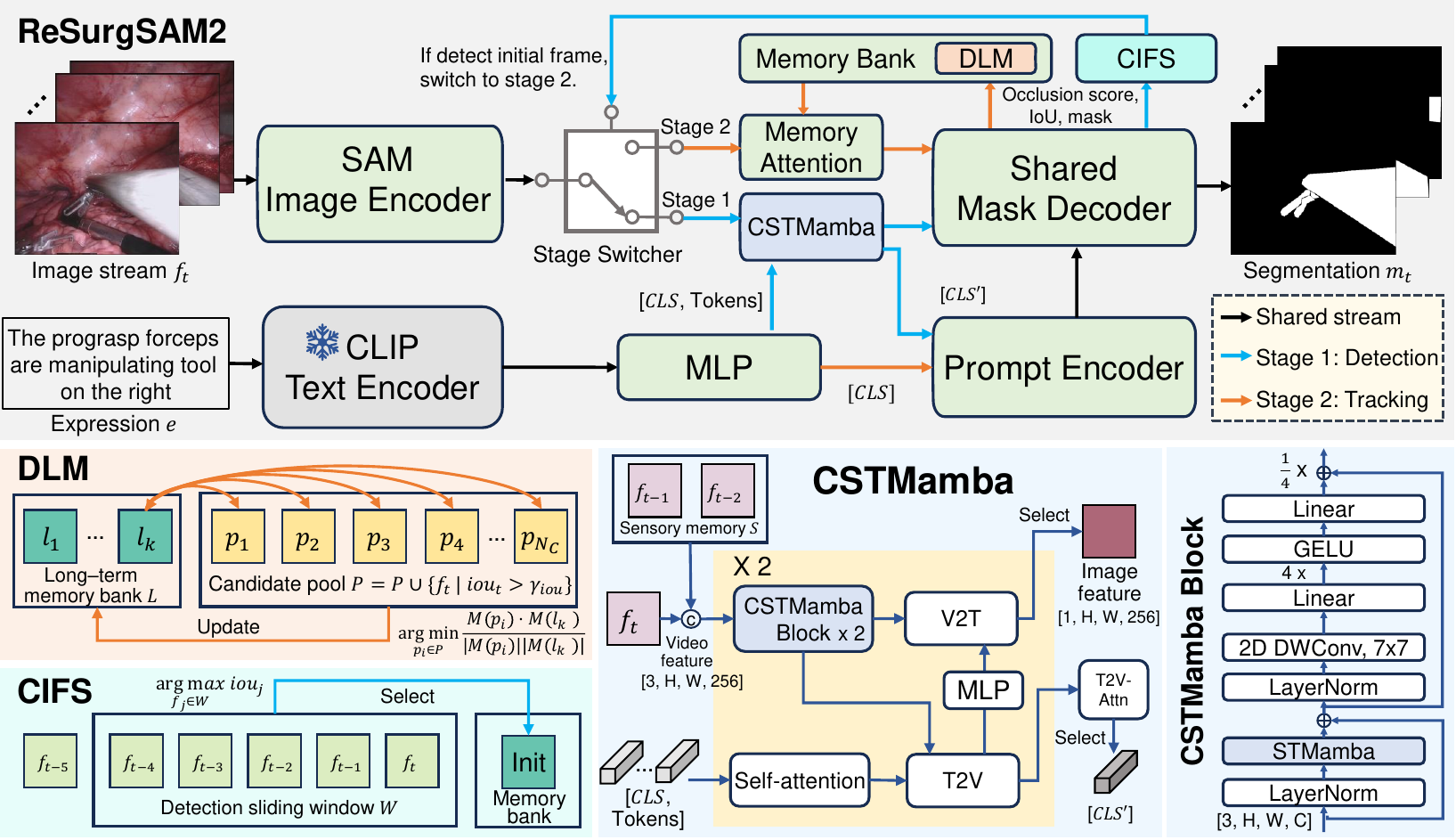}
    \caption{Overview of ReSurgSAM2. The model begins with the text-referred target detection using CSTMamba to provide credible frames for selection using CIFS. Upon detecting the initial frame, CIFS activates the tracking stage, in which DLM offers diverse and reliable memory for consistent long-term tracking.}
    \label{fig:architecture}
\end{figure}

To overcome the above limitations, we propose ReSurgSAM2, a specialized two-stage framework for surgical applications that seamlessly integrates text-referred target detection with tracking, as illustrated in Fig.~\ref{fig:architecture}.
Given the $t$-th frame $f_t \in \mathbb{R}^{3 \times H \times W}$ from an image stream and its linguistic expression $e$, we separately employ SAM image encoder and the frozen CLIP text encoder with a trainable MLP to extract features, with our goal of obtaining the segmentation mask $m_t\in \mathbb{R}^{H\times W}$ for the text-referred target. 
In the first stage, our model enhances detection reliability by utilizing CSTMamba and a mask decoder to generate a high-fidelity segmentation mask. 
The scores of the mask are subsequently fed into the CIFS, which performs credible frame selection prepared for the tracking phase to mitigate error accumulation.
Upon identification of the optimal initial frame by CIFS, the model switches to the tracking stage, where the prompt encoder accepts the $CLS$ token from text features, and the model ensures reliable and consistent object tracking throughout the video 
by integrating vanilla short-term memory with long-term memory using the DLM.

\subsection{Identify the Reliable Initial Frame with CSTMamba and CIFS}
\label{sec:cstmamba}
\noindent \textbf{Cross-Modal Spatial-Temporal Mamba (CSTMamba):}
Referring segmentation using a single frame often yields a suboptimal result, and propagating this result as an initial reference for tracking potentially causes error accumulation. Therefore, precise target detection is crucial for our two-stage framework.
While transformer-based approaches can enhance segmentation by leveraging self-attention, this mechanism suffers from quadratic complexity~\cite{tay2022efficient}, limiting the real-time surgical applications.
Recently, Mamba~\cite{mamba} has emerged as a promising alternative with linear complexity and selective information propagation. 
STMamba~\cite{vivim} leverages these advantages for video segmentation; however, it lacks cross-modal capabilities for RVOS and is constrained by Mamba's linear scanning mechanism, which restricts its ability to capture fine-grained pixel-level information.

To address these limitations, we propose the CSTMamba, which integrates a sensory memory bank, the CSTMamba block, cross-modal attention mechanism to facilitate comprehensive cross-modal spatial-temporal modeling.
To enable temporal interaction, the sensory memory bank $S$ stores the most recent two frame features extracted from the image encoder.
Then, the CSTMamba takes the language features and video features from the current frame and sensory memory as input for cross-model spatial-temporal modeling, and outputs the fused image feature $E(f_t)'$ and $CLS'$ token.
The video input can be written as $[E(f_t),E(f_{t-1}),E(f_{t-2})]\in \mathbb{R}^{3 \times H' \times W'\times 256} $, where $E(\cdot)$ is the image encoder and $[\cdot]$ is concatenation, $H'$ and $W'$ are the feature height and width.

The CSTMamba block, designed for comprehensive spatial-temporal modeling, is illustrated in Fig.~\ref{fig:architecture}.
It integrates STMamba with a $7\times7$ 2D depth-wise convolution (DWConv) to capture global features with selective scanning and fine-grained local features with an expanded receptive field. 
Building upon this foundation, an inverted bottleneck~\cite{sandler2018mobilenetv2} expands the MLP block to four times the input dimension, effectively enhancing feature representation by utilizing the enlarged middle layer to improve spatial-temporal interactions. 
Additionally, bidirectional cross-modal attention mechanisms—text-to-vision (T2V) and vision-to-text (V2T)~\cite{vaswani2017attention}—facilitate cross-modal interactions. 
The CSTMamba processes visual-temporal information and cross-modal signals simultaneously, generating enriched spatial-temporal features and a fused $CLS'$ token with strong cross-modal representations for final mask prediction.

\noindent \textbf{Credible Initial Frame Selection (CIFS):}
When automatically selecting an initial reference frame for tracking, it is essential to identify a highly reliable frame to mitigate error accumulation~\cite{liu2022learning}.
However, in surgical environments, the small inter-class similarity among instruments and among tissues increases the risk of false detections, leading to unreliable segmentation.
To address this challenge, we implement CIFS that requires the model to predict the object as present with high confidence based on IoU score and occlusion score across $N_w$ consecutive frames before frame selection begins. Before selection, it uses the sliding window to detect the qualified frames, which can be formulated as:
\begin{equation}
\medmath{
    W=\{f_j \mid j \in [t-N_w+1, t] \land iou_j > \delta_{iou} \land sigmoid(o_j) > \delta_{o}\}, 
}
\end{equation}
where $W$ is the sliding window, $o_j$ and $iou_j$ are the predicted occlusion score and IoU score for the $t$-th frame, $\delta_{iou}$ and $\delta_{o}$ are their respective thresholds, and $sigmoid(\cdot)$  maps the score to range $[0,1]$. 
Upon $|W|=N_w$, the one with the highest IoU score is selected as the initial reference among these $N_w$ qualifying frames.
With the selection of an optimal and credible initial frame, ReSurgSAM2 enters its second stage, where it performs robust tracking by leveraging this reference frame to propagate predictions throughout the remaining sequence, ensuring both semantic fidelity and temporal consistency.

\subsection{Diversity-driven Long-term Memory}
\label{sec:dlm}
In surgical environments, videos typically have long durations with dynamic scene variations and instrument movement. However, SAM2's greedy strategy of selecting only the most recent frames as memory hinders effective long-term tracking, leading to redundancy and potential viewpoint overfitting, thereby limiting its ability to capture anatomical changes.
To overcome this challenge, we propose the DLM mechanism that enhances SAM2's vanilla memory bank by strategically selecting frames from a candidate pool. 
This mechanism enriches the memory bank by using the candidate pool to extend the temporal coverage range and collecting the frames that capture diverse spatial-temporal information to mitigate viewpoint overfitting. Additionally, it ensures the inclusion of high-confidence frames to minimize error accumulation.

The proposed DLM mechanism updates the candidate pool in inference as:
\begin{equation}
    \medmath{P = P \cup \{f_t \mid iou_{t} > \gamma_{iou}\}},
\end{equation}
where $P$ is the candidate pool, $f_t$ is the $t$-th frame, ${iou}_{t}$ is its predicted IoU score, and $\gamma_{iou}$ is a confidence threshold. 
Each element in $P$ is indexed by $p_i$, where $p_i$ refers to the $i$-th candidate.  
This mechanism selects high-confidence frames as memory candidates for mitigating error propagation.
When the candidate pool reaches its capacity $N_p$, we store the most diverse candidate based on its cosine similarity to the latest long-term memory frame:
\begin{equation}
\medmath{p^* = \mathop{\arg\min}_{p_i \in P} \frac{M(p_i) \cdot M(l_{k})}{|M(p_i)| |M(l_{k})|}},
\end{equation}
where $M(\cdot)$ is the memory encoder, $p_i$ is $i$-th candidate frame, $l_{k}$ is the latest frame in long-term memory bank $L$, and $p^*$ is the selected frame. 
After selection, the pool $P$ is cleared to extend the temporal coverage range of the long-term memory, and the updated memory bank is utilized for the ${(t+1)}$-th frame's memory attention mechanism.
To enhance efficiency, we maintain a queue with capacity $N_l$ for long-term memory, keeping the initial frame permanently in the long-term memory.
By concatenating SAM2's vanilla short-term memory with our long-term memory using the DLM mechanism, ReSurgSAM2 maintains a memory bank that is reliable and diverse, boosting consistent long-term tracking. 

\section{Experiment}
\begin{table}[tbp]
  \scriptsize
  \centering
  \caption{Dataset statistics for Ref-EndoVis17 and Ref-EndoVis18.}
    \begin{tabular}{l|cccc|cccc}
    \hline
    \multirow{2}[0]{*}{Dataset}      & \multicolumn{4}{c|}{Training}  & \multicolumn{4}{c}{Testing} \\
    ~ & Sequence & Frame & Object & Pair & Sequence & Frame & Object & Pair \\
    \hline
    Ref-EndoVis17(tool) & 7     & 2100  & 20    & 4873  & 3     & 900   & 10    & 2265 \\
    Ref-EndoVis18(tool) & 11 & 1639 & 34    & 3787  & 4 & 596 & 15    & 1384 \\
    Ref-EndoVis18(tissue) &  11   &  1639 & 25    & 2995  & 4 & 596 & 7     & 807 \\
    \hline
    \end{tabular}%
  \label{tab:dataset} 
\end{table}%

\begin{table}[bp]
    \scriptsize
    \centering
    \caption{
    Quantitative comparison with state-of-the-art methods.
    }
    \label{tab:comparison} 
    \renewcommand{\arraystretch}{1.2}
    \begin{tabular}{m{2.1cm} | c | p{0.8cm}<{\centering} p{0.8cm}<{\centering} p{0.8cm}<{\centering} | p{0.8cm}<{\centering} p{0.8cm}<{\centering} p{0.8cm}<{\centering}| p{0.8cm}<{\centering} p{0.8cm}<{\centering} p{0.8cm}<{\centering}| c}
    \hline
    \multirow{2}*{Method}& \multirow{2}*{Setting}  & \multicolumn{3}{c|}{Ref-EndoVis17(tool)}  &\multicolumn{3}{c|}{Ref-EndoVis18(tool)}&\multicolumn{3}{c|}{Ref-Endovis18(tissue)} & \multirow{2}*{FPS}\\
    ~ & ~ & $\mathcal{J}$\&$\mathcal{F}$  & $\mathcal{J}$ & $\mathcal{F}$ & $\mathcal{J}$\&$\mathcal{F}$  & $\mathcal{J}$ & $\mathcal{F}$ & $\mathcal{J}$\&$\mathcal{F}$  & $\mathcal{J}$ & $\mathcal{F}$ & ~ \\ 
    \hline
    ReferFormer~\cite{wu2022language} & Offline & 62.41 & 62.28 & 62.55 & 71.09 & 70.96 & 71.23 & 61.84 & 69.9 & 53.78 & 42.3\\
    MUTR~\cite{yan2024referred} & Offline & 60.97 & 60.76 & 61.18 & 67.56 & 67.79 & 67.33 & 63.53 & 71.48 & 55.58 & 32.3\\
    RSVIS~\cite{wang2024video} & Online & 61.22 & 61.37 & 61.07 & 68.35 & 68.55 & 68.15 & 65.69 & 72.91 & 58.47 & 22.1\\
    OnlineRefer~\cite{wu2023onlinerefer} & Online & 60.32 & 60.29 & 60.34 & 72.19 & 71.88 & 72.50 & 70.56 & 77.58 & 63.55 & 25.6\\
    RefSAM~\cite{li2024refsam} & Online & 63.56 & 63.77 & 63.35 & 72.86 & 73.40 & 72.31 & 71.90 & 77.66 & 66.14 & 25.4\\
    \hline
    ReSurgSAM2 & Online & \textbf{77.73} & \textbf{77.77} & \textbf{77.69} & \textbf{80.62} & \textbf{80.94} & \textbf{80.31} & \textbf{75.09} & \textbf{80.93} & \textbf{69.25} & \textbf{61.2}\\
    \hline
    \end{tabular}
\end{table}
\subsection{Dataset and Implementation Details}
The experiments were conducted with Ref-EndoVis17 and Ref-EndoVis17 building upon EndoVis17~\cite{ref_Allan2019EndoVis17}, EndoVis18 dataset~\cite{ref_Allan2020EndoVis18} and RSVIS~\cite{wang2024video}.
The EndoVis17 comprises 3000 frames across 10 sequences, including eight training sequences, eight test sequences from identical scenes, and two additional test sequences, with instrument labels.
The EndoVis18 dataset consists of 15 sequences with comprehensive scene segmentation annotations.
As introduced in RSVIS, both datasets were reannotated with consistent instance-specific labels to facilitate RVOS. 
Building upon RSVIS~\cite{wang2024video}, we performed meticulous refinement to address inconsistencies and omissions in instrument labeling that would otherwise compromise experimental validity.
We further enriched the datasets by incorporating tissue-specific annotations from EndoVis18, including kidney parenchyma, covered kidney, and small intestine.
For Ref-EndoVis17, we merged sequences from the identical scenes to prevent cross-contamination between training and test sets, with sequences 2, 5, and 6 designated as the test set. 
Following RSVIS~\cite{wang2024video}, we allocated sequences 2, 5, 9, and 15 as the test set for Ref-EndoVis18.
This division ensures balanced object distribution across training and testing. 
Table~\ref{tab:dataset} presents statistics for both datasets, with "pair" denoting the text-mask pair.

ReSurgSAM2 employs the Hiera-small backbone~\cite{ryali2023hiera} initialized with SAM2 pre-trained weights~\cite{sam2}, with an input size of $512$. 
During training, we follow SAM2 to conduct prompt segmentation by loading three frames for text-referred object detection, followed by seven frames for tracking. 
The model is trained for 30 epochs utilizing the same training strategies as SAM2. 
For inference, unlike RSVIS, we generate text expressions at the first appearance of each object to accommodate the whole surgical video.
The hyperparameters are set as follows: $\delta_{o}=0.9$, $\delta_{iou}=0.7$, $\gamma_{iou}=0.95$, $N_w=5$, $N_p=5$ and $N_l=4$.
Unlike semantic segmentation using challenge IoU~\cite{ref_Allan2019EndoVis17,ref_Allan2020EndoVis18}, RVOS tracks the object across the whole video, including when it is occluded.
Therefore, for evaluation metrics, we adopt $\mathcal{J}$ and $\mathcal{F}$~\cite{perazzi2016benchmark} for accuracy, where $\mathcal{J}$ assesses region accuracy and $\mathcal{F}$ evaluates boundary accuracy following RVOS~\cite{khoreva2019video}, with $\mathcal{J}\&\mathcal{F}$ representing their mean, and the frame-per-second (FPS) for efficiency.
All metrics are higher-is-better. All experiments utilized the same training data on an NVIDIA A6000 GPU.

\begin{figure}[tbp]
    \centering
    \includegraphics[width=\textwidth]{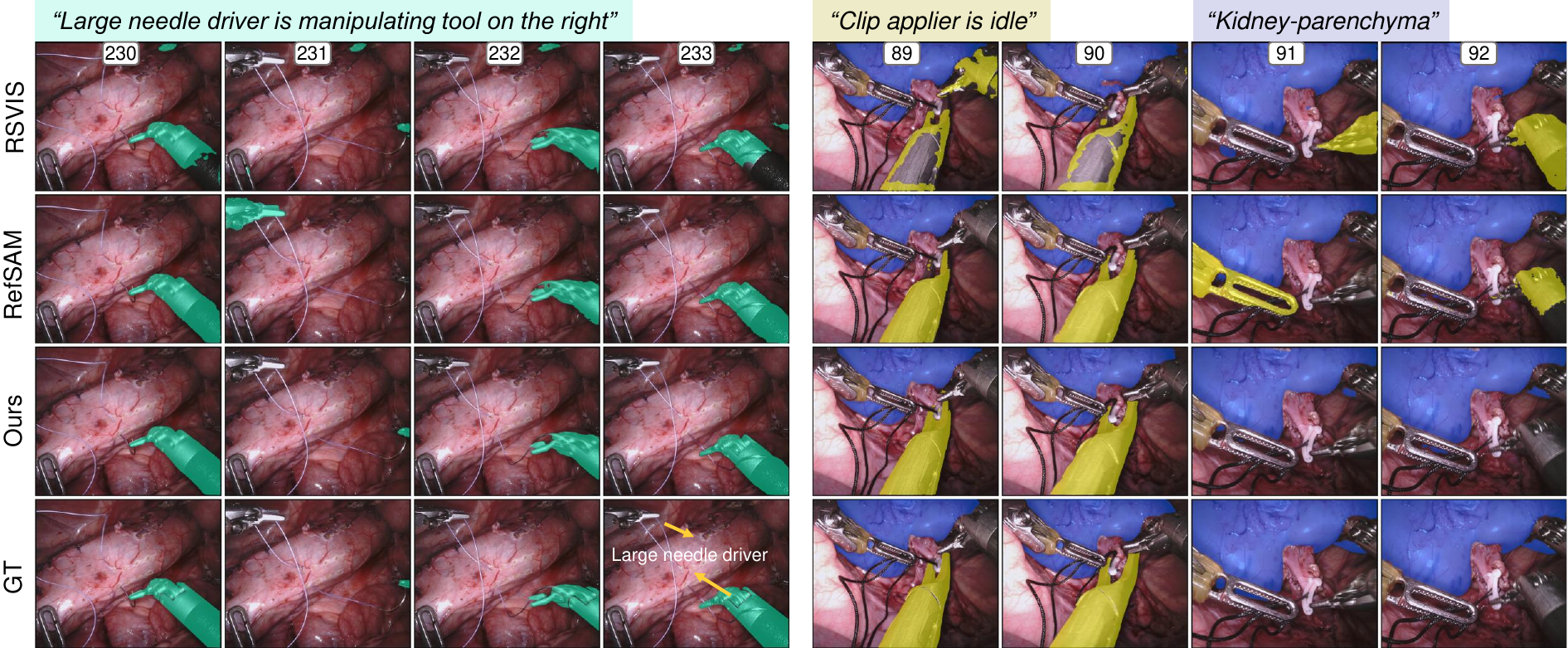}
    \caption{Qualitative comparison in Ref-EndoVis17(left) and Ref-EndoVis18(right).}
    \label{fig:comparison}
\end{figure}

\subsection{Comparison and Ablation Study}
\noindent \textbf{Comparison:}
To validate the effectiveness of ReSurgSAM2, we conducted comprehensive comparisons against state-of-the-art methods summarized in Table~\ref{tab:comparison}. 
The comparison methods include offline methods (ReferFormer and MUTR) and online methods (RSVIS, OnlineRefer, and RefSAM).
Offline methods achieve stable performance by processing 64 frames concurrently during inference, thus reducing false detections.
In contrast, RSVIS only relies on short-term information, leading to suboptimal long-term tracking.
While OnlineRefer and RefSAM exhibit modest long-term tracking capabilities by query propagation, the performance remains suboptimal on long-range sequences in Ref-EndoVis17.
In comparison, ReSurgSAM2 shows superior cross-modal and long-term tracking capabilities in RVOS, with substantial $\mathcal{J}\&\mathcal{F}$ improvements: 14.17 on Ref-EndoVis17, 7.76 on Ref-EndoVis18 tool, and 3.19 on Ref-EndoVis18 tissue datasets.

The qualitative comparison with RSVIS and RefSAM is illustrated in Fig.~\ref{fig:comparison}. 
RSVIS lacks robust instrument discrimination in complex scenarios, causing incomplete segmentation.
Both RSVIS and RefSAM exhibit limited tracking stability during rapid object movements and scene variations, due to their limited long-term modeling.
In contrast, ReSurgSAM2, equipped with a robust initialization and diverse long-term memory, performs reliable and consistent tracking.

\noindent \textbf{Ablation Study:}
We conduct comprehensive ablation studies on Ref-EndoVis17 to evaluate the effectiveness of each proposed component, with results presented in Table~\ref{tab:ablation}. 
For ablation, the two-stage RVOS framework uses the first detected frame with $iou_t>0.7$ and $sigmoid(o_t) > 0.9$ as a reference to activate tracking of stage 2.
This design helps capture short-term temporal dependencies, leading to a $\mathcal{J}\&\mathcal{F}$ gain of 2.64. 
Integrating CSTMamba strengthens spatial-temporal referring segmentation to generate a more reliable reference for tracking, achieving a 4.77 improvement in $\mathcal{J}\&\mathcal{F}$.
Furthermore, the CIFS strategy further enhances $\mathcal{J}\&\mathcal{F}$ by 6.14, as it selects a more reliable reference as memory, mitigating error accumulation.
DLM further enhances vanilla memory mechanism with long-term temporal modeling by maintaining a diverse and long-range memory bank, contributing a 3.03 boost in $\mathcal{J}\&\mathcal{F}$.
To validate DLM's effectiveness, we explored various memory bank designs: an extended short-term memory based on the vanilla memory bank, and a long-term memory with interval sampling every five frames (storing three frames each), as shown in Table~\ref{tab:dlm}.
By integrating all proposed components, ReSurgSAM2 ultimately achieves 77.73\ in $\mathcal{J}\&\mathcal{F}$ while maintaining real-time performance at 61.2 FPS.

\vspace{8pt}
\noindent
\begin{minipage}{.62\textwidth}
    \centering
    \scriptsize
    \renewcommand{\arraystretch}{1.2}
    \begin{tabular}{c|c|c|c|c c c | c}
    \hline
    Stage 2 & CSTMamba & CIFS & DLM & $\mathcal{J}$\&$\mathcal{F}$ & $\mathcal{J}$ & $\mathcal{F}$ & FPS\\
    \hline
    & & &                                & 61.15 & 61.46 & 60.84 & \textbf{70.1} \\
    $\surd$ &         &         &         & 63.79 & 63.77 & 63.82 & 68.2 \\
    $\surd$ & $\surd$ &         &         & 68.56 & 68.51 & 68.61 & 67.5 \\
    $\surd$ & $\surd$ & $\surd$ &         & 74.70 & 74.67 & 74.72 & 63.1 \\
    $\surd$ & $\surd$ & $\surd$ & $\surd$ & \textbf{77.73} & \textbf{77.77} & \textbf{77.69} & 61.2 \\
    \hline
    \end{tabular}
    \captionof{table}{Ablation studies on Ref-Endovis17.}
    \label{tab:ablation}
\end{minipage}
\hfill
\begin{minipage}{0.37\textwidth}
    \centering
    \scriptsize
    \renewcommand{\arraystretch}{1.2}
    \setlength\tabcolsep{2pt}
    \begin{tabular}{c| c c c}
    \hline 
    Method & $\mathcal{J}$\&$\mathcal{F}$ & $\mathcal{J}$ & $\mathcal{F}$ \\
    \hline
    Vanilla & 74.70 & 74.67 & 74.72 \\
    Extended &  74.68 & 74.64 & 74.72 \\
    Interval &  75.32 & 75.27 & 75.37\\
    DLM & \textbf{77.73} & \textbf{77.77} & \textbf{77.69}\\
    \hline
    \end{tabular}
    \captionof{table}{Ablation on memory bank design.}
    \label{tab:dlm}
\end{minipage}

\vspace{-5pt}
\section{Conclusion}
This paper presents ReSurgSAM2, a two-stage framework via credible long-term tracking for surgical referring segmentation.
While existing methods are limited in efficiency and long-term tracking, our approach addresses these limitations through reliable initial frame identification and a long-term memory mechanism building upon SAM2. 
Extensive experiments on surgical datasets demonstrate that ReSurgSAM2 significantly outperforms existing methods, offering a practical and efficient solution for real-time surgical video analysis.

\bibliographystyle{splncs04}
\bibliography{mybibliography}

\end{document}